\documentclass[sigconf]{acmart}

\AtBeginDocument{%
	}
\usepackage{multirow}
\usepackage{graphicx}
\usepackage{balance} 
\usepackage{pifont}
\usepackage{wasysym}
\usepackage{utfsym}
\usepackage{fontawesome}
\usepackage{fancyhdr}
\renewcommand\footnotetextcopyrightpermission[1]{}

\begin{document}
\title{MindDiffuser:\\Controlled Image Reconstruction from Human Brain Activity with Semantic and Structural Diffusion}
\bibliographystyle{unsrt}

\author{Yizhuo Lu}
\authornote{Both authors contributed equally to this research.}
\affiliation{%
	\institution{Laboratory of Brain Atlas and Brain-Inspired Intelligence, State Key Laboratory of Multimodal Artificial Intelligence Systems, CASIA}
	\institution{School of Future Technology, University of Chinese Academy of Sciences, Beijing, China}
	\city{ }
	\country{}
}
\email{luyizhuo2023@ia.ac.cn}

\author{Changde Du}
\authornotemark[1]
\affiliation{%
	\institution{Laboratory of Brain Atlas and Brain-Inspired Intelligence, State Key Laboratory of Multimodal Artificial Intelligence Systems, CASIA}
	\city{Beijing}
	\country{China}
}

\email{changde.du@ia.ac.cn}

\author{Qiongyi zhou}
\affiliation{%
	\institution{Laboratory of Brain Atlas and Brain-Inspired Intelligence, State Key Laboratory of Multimodal Artificial Intelligence Systems, CASIA}
	\institution{University of Chinese Academy of Sciences, Beijing, China}
	\city{ }
	\country{}
}
\email{zhouqiongyi2018@ia.ac.cn}

\author{Dianpeng Wang}
\affiliation{%
	\institution{	School of Mathematics and Statistics}
	\institution{Beijing Institute of Technology}
	\city{Beijing}
	\country{China}}
\email{wdp@bit.edu.cn}

\author{Huiguang He}
\authornote{corresponding author.}
\affiliation{%
	\institution{Laboratory of Brain Atlas and Brain-Inspired Intelligence, State Key Laboratory of Multimodal Artificial Intelligence Systems, CASIA}
	\institution{University of Chinese Academy of Sciences, Beijing, China}
	\city{ }
	\country{}
}
\email{huiguang.he@ia.ac.cn}


	\begin{teaserfigure}
	\centering
	\includegraphics[width=0.9\textwidth]{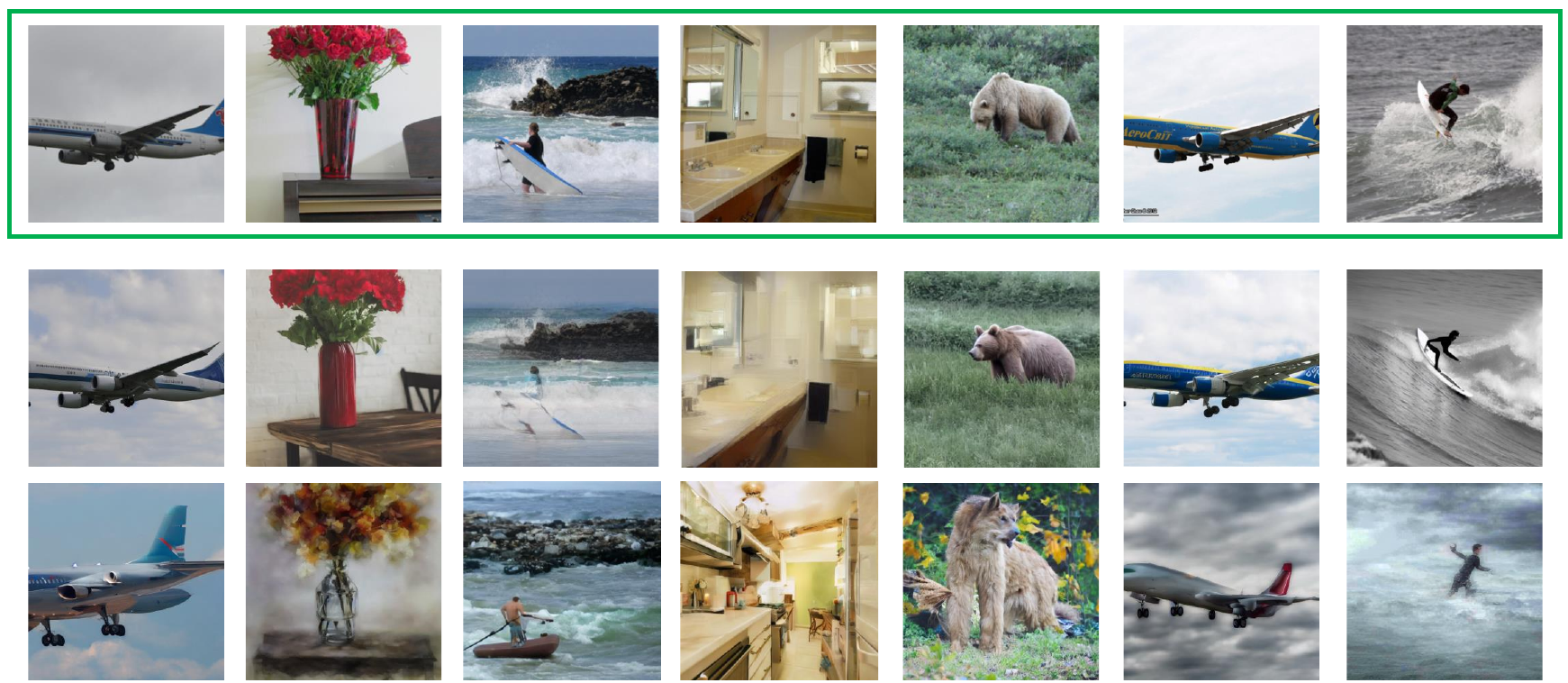}
	\caption{An overview of images reconstructed by MindDiffuser. The top row (green box) represents the image stimuli, the second row represents the upper bound of the model's reconstruction ability, using true feature extracted from image stimuli. The last row shows our results with good alignment of both semantic and structural information to the stimuli.}
	\Description{Overview of our results.}
	\label{fig:result-overview}
	\thispagestyle{empty}
	\pagestyle{empty}
\end{teaserfigure}

\begin{abstract}
	\thispagestyle{empty}
	\pagestyle{empty}
	Reconstructing visual stimuli from brain recordings has been a meaningful and challenging task. Especially, the achievement of precise and controllable image reconstruction bears great significance in propelling the progress and utilization of brain-computer interfaces. Despite the advancements in complex image reconstruction techniques, the challenge persists in achieving a cohesive alignment of both semantic (concepts and objects) and structure (position, orientation, and size) with the image stimuli. To address the aforementioned issue, we propose a two-stage image reconstruction model called MindDiffuser\footnote{ The code has been released at: \href{https://github.com/ReedOnePeck/MindDiffuser}{https://github.com/ReedOnePeck/MindDiffuser}}. In Stage 1, the VQ-VAE latent representations and the CLIP text embeddings decoded from fMRI are put into  Stable Diffusion, which yields a preliminary image that contains semantic information. In Stage 2, we utilize the CLIP visual feature decoded from fMRI as supervisory information, and continually adjust the two feature vectors decoded in Stage 1 through backpropagation to align the structural information. The results of both qualitative and quantitative analyses demonstrate that our model has surpassed the current state-of-the-art models  on Natural Scenes Dataset (NSD). The subsequent experimental findings corroborate the neurobiological plausibility of the model, as evidenced by the interpretability of the multimodal feature employed, which align with the corresponding brain responses. 
\end{abstract}


\keywords{Probabilistic diffusion model, fMRI, Controlled Image Reconstruction, Brain-computer Interface (BCI)}

\maketitle
\thispagestyle{empty}
\pagestyle{empty}

\section{Introduction}  

The human visual system possesses the exceptional ability to efficiently and robustly perceive and comprehend complex visual stimuli in the real world, which is unparalleled by current artificial intelligence systems.  Understanding these brain activities and reconstructing \cite{rakhimberdina2021natural}  the corresponding stimuli is a critical step towards the ultimate goal of deciphering the workings of the human brain, despite its immense difficulty.  With the advancement of sophisticated image generation methods and the increase in the volume of neuroimaging data, researchers are increasingly focusing on image reconstruction.

Recently, researches have revealed that deep learning frameworks exhibit a certain level of consistency with the hierarchical encoding-decoding process of the human visual system \cite{pinto2009high, krizhevsky2009learning, schrimpf2018brain}. As a result, numerous studies have extensively employed deep neural networks (DNN) for reconstructing natural images. Based on the structure of the previous image reconstruction models, we categorize them into  {\bfseries optimized models} and {\bfseries generative models}. The optimized model is represented by DGN \cite{shen2019deep} proposed by Shen et al., which utilizes image feature extracted from a DNN as a constraint, and optimizes the latent space of the image generator to achieve similarity with the decoded DNN feature. While this method allows for alignment of the structural information of the reconstructed images with the corresponding ones in pixel space, the absence of image priori in latent space means that optimization starting from Gaussian noise can result in indistinct outcomes and a lack of clear semantic information. The generative reconstruction models involve decoding fMRI into the latent space of models such as VAE \cite{ilic2019auto}, GAN \cite{goodfellow2020generative}, and Diffusion model \cite{song2019generative}, and leveraging their powerful generation capabilities to reconstruct images that are semantically similar to the original. While this paradigm enables rapid generation of realistic and semantically rich reconstruction images, the outcomes are always lacking in control over structural information.

Building upon the preceding discussions, we make the following contributions in this work: \\
(1) We present an image reconstruction model (MindDiffuser) that integrates the strengths of the two aforementioned paradigms and effectively addresses their respective limitations, resulting in semantically similar and structurally aligned reconstruction outcomes. A series of detailed quantitative comparisons demonstrate that our model has surpassed the current state-of-the-art  models.\\
(2) Our proposed two-stage image reconstruction model leverages semantic feature incorporating and structural feature aligning to achieve a remarkably powerful image reconstruction capability, as evidenced by the upper bound showcased in Figure \ref{fig:result-overview}. Additionally, our model can serve as a module and be combined with future advancements in more effective brain signal decoders to achieve higher-quality visual reconstruction results.\\
(3) Our experiments demonstrate that MindDiffuser can adapt to inter-subject variability without additional modifications. Furthermore, the visualization of feature decoding precosses provide evidence of the model's rationality and interpretability in neuroscience.

\section{Related works}
\label{related_works}
\subsection{Neural decoding and image reconstruction models}
\label{subsection:related_works_one}
Building upon the pioneering work of Haxby \cite{haxby2001distributed}, a multitude of neural decoding tasks with significant guiding implications have surfaced in recent decades. These tasks can be categorized into stimulus classification \cite{haxby2001distributed,van2010efficient,damarla2013decoding,yargholi2016brain,du2022decoding}, stimulus identification \cite{haynes2006decoding,kay2008identifying,naselaris2009bayesian,horikawa2017generic}, and stimulus reconstruction \cite{nishimoto2011reconstructing, 10.3389/fncom.2019.00021, beliy2019voxels, gaziv2020self, fang2020reconstructing, ren2021reconstructing} based on their decoding objectives. Among them, stimuli reconstruction is the most alluring and demanding, and we focus on this area in the study.

Previous image reconstruction techniques utilized linear regression models to fit fMRI with manually defined image features \cite{kay2008naselaris,naselaris2009bayesian,fujiwara2013modular}. The outcomes were blurry, and the features relied heavily on manual configuration.
With the advent of deep learning, the use of DNNs in this domain has become more prevalent. Beliy et al. \cite{beliy2019voxels} and Gaziv et al. \cite{gaziv2022self} employed semi-supervised learning \cite{chapelle2009semi} to train an Encoder-Decoder structure to reconstruct images, solving the problem of stimulus-fMRI pairs deficiency. However, the results of these models do not possess distinguishable semantic information.
Du et al. \cite{du2018reconstructing} introduced a multi-view reconstruction model that accounts for the statistical correlation between fMRI signals and the corresponding stimuli. Du et al. \cite{du2020conditional,du2020structured} enhances the quality of partial natural images and face reconstructions through the use of a DNN and a matrix variable Gaussian priori, employing multi-task transfer learning.
Chen et al. \cite{chen2022seeing} pre-trained fMRI data using a method similar to MAE \cite{he2022masked}, and fine-tuned the LDM \cite{rombach2022high} to obtain reconstructed images.
Ozcelik et al. \cite{ozcelik2022reconstruction} and Gu et al. \cite{gu2022decoding} employed a self-supervised model for image instance feature extraction, followed by iterative optimization of noise and dense information using backpropagation. Subsequently, the three feature vectors were mapped from fMRI and put into IC-GAN \cite{casanova2021instance} for image reconstruction.

Since the introduction of Contrastive Language-Image Pre-training (CLIP) \cite{radford2021learning}, semantic information from text has been leveraged for reconstructing complex natural images. Lin et al. \cite{lin2022mind} trained a mapping model using contrastive and adversarial learning loss to align fMRI with CLIP latent representations. The mapped fMRI was then fed into StyleGAN2 \cite{karras2020analyzing} during the generation stage. Takagi et al. \cite{takagi2022high} achieved close-to-original reconstruction results by mapping fMRI to the text feature $c$ and image feature $z$ of Stable Diffusion \cite{rombach2022high}.
The most relevant work within our context is by Takagi et al. However, since no additional constraints are imposed on the generator, their resulting reconstructions lack precision in detail such as location, size, and shape.
\subsection{CLIP for image generation}
CLIP \cite{radford2021learning} utilizes image-text contrastive learning to endow its representation space with rich semantic information, which has been widely utilized to guide downstream image generation tasks. Here, We describe the following works according to the different roles that CLIP plays in downstream tasks.

{\bfseries Feed-forward image generation based on CLIP.} DALL$\cdot$E 2 \cite{ramesh2022hierarchical} employs a prior model to transform CLIP text embeddings into corresponding image feature, which are subsequently utilized by a diffusion model to generate images. Stable Diffusion \cite{rombach2022high} utilizes CLIP text embeddings to guide the denoised process in image generation, thereby ensuring high quality output.

Besides its direct application in guiding image generation, the feature extracted by CLIP can also serve as supervisory signals for continuously refining the latent vectors of the image generator via backpropagation. This technique enables fine-grained and personalized image generation and editing.

{\bfseries Backward image optimization supervised by CLIP.}  StyleCLIP \cite{patashnik2021styleclip} begins by obtaining the latent vectors for a given image through GAN Inversion \cite{xia2022gan}. It then optimizes the latent vectors under the supervision of input text's CLIP feature, resulting in an image that aligns with the given text. CLIPasso \cite{vinker2022clipasso} extracts the structure and semantic information of the original image and the generated stick figure image, whose L2 distance is used as a loss function, and then optimizes the stroke parameters until convergence through backpropagation. Similar to CLIPasso, our MindDiffuser uses the low-level image feature extracted from CLIP to constrain the structural information of the reconstructed images.

The prior works leveraging the CLIP representation space mentioned in \ref{subsection:related_works_one} (Lin \cite{lin2022mind}, Takagi \cite{takagi2022high}) utilize the decoded CLIP feature to guide the image reconstruction process directly. To the best of our knowledge, MindDiffuser is the first approach to leverage CLIP feature as supervisory information for achieving fine-grained and controlled image reconstruction.

\section{Method}
\subsection{Overview}

\begin{figure*}[htbp!]
	\centering
	\includegraphics[width=1.0\linewidth]{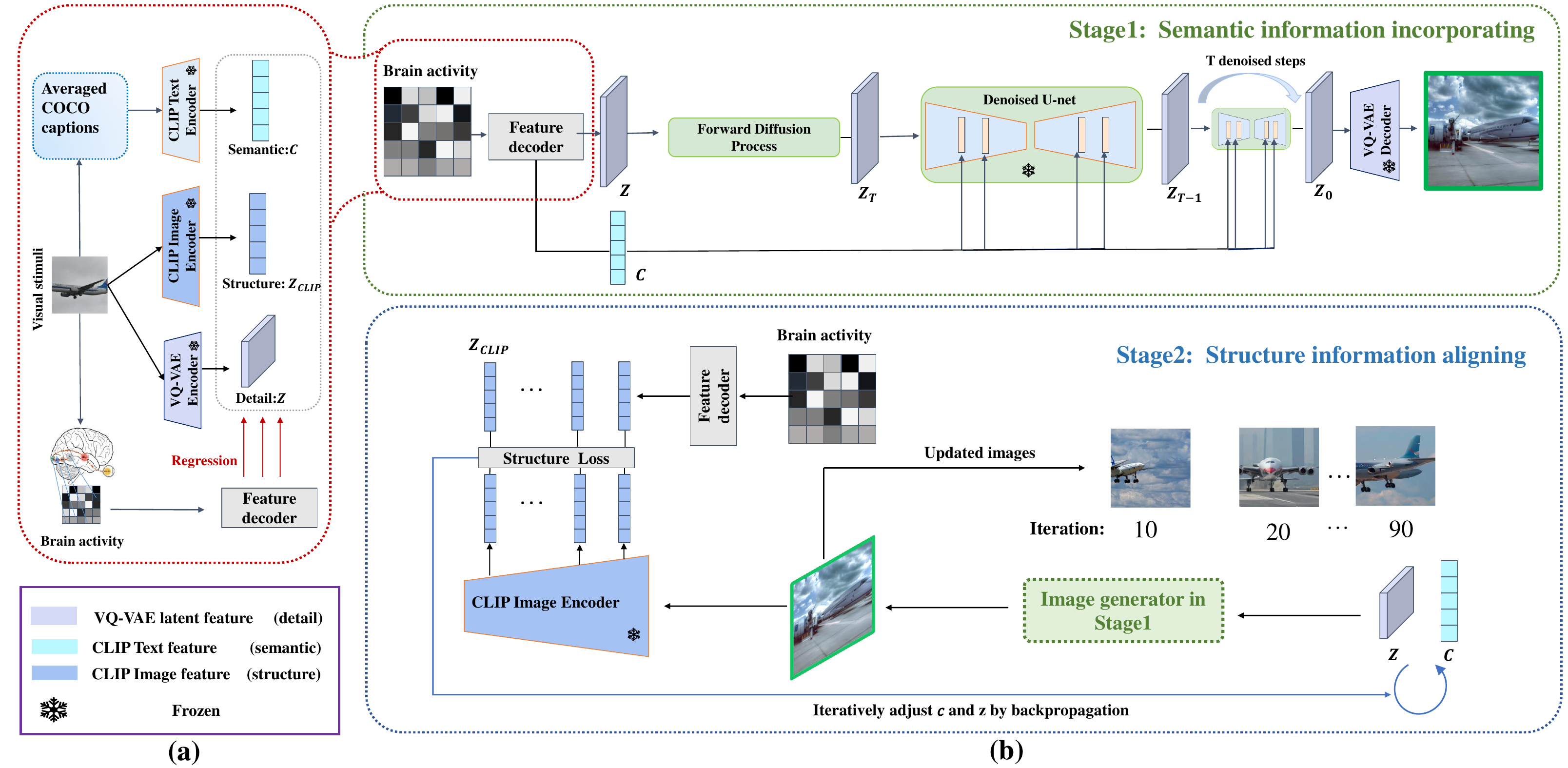}
	\caption{Schematic diagram of MindDiffuser. (a) Decoders are trained to fit fMRI with averaged CLIP text embeddings $c$, CLIP image feature $Z_{CLIP}^i$, and VQ-VAE latent feature $z$. (b) The two-stage image reconstruction process. In stage 1, an initial reconstructed image is generated using the decoded CLIP text feature $c$ and VQ-VAE latent feature $z$. In stage 2, the decoded CLIP image feature is used as a constraint to iteratively adjust $c$ and $z$ until the final reconstruction result matches the original image in terms of both semantic and structure.}
	\label{fig:overview}
\end{figure*}

In this section, we present {\bfseries MindDiffuser}, a two-stage model for controlled image reconstruction, as shown in Figure \ref{fig:overview}. Briefly, in stage 1, we decode fMRI into CLIP text embeddings $c$ and visual feature $z$ in the VQ-VAE latent space. This enables the initial reconstructed images generated by Stable Diffusion to contain semantic information and fine-grained detail, thereby interpreting {\bfseries ``What is contained in the image ?''} In stage 2, we decode fMRI into the low-level visual feature of CLIP, and continuously optimize $c$ and $z$ in stage 1 by back-propagation, so that the generated images approximate the ground truth in the CLIP embbeding space, thus achieving control over the structural information and deciphering {\bfseries ``Where are the objects in the image ?''}

\subsection{Stage 1: Semantic information incorporating}

Suppose that $Y \in R^{D_y \times N}$ and $X \in R^{D_x \times N}$  denote the visual stimuli and its corresponding fMRI activity patterns in the training set, respectively. And $c \in R^{D_c \times N}$, $z \in R^{D_z \times N}$ and $Z_{CLIP}^i \in R^{D_i \times N}$ denote CLIP text embeddings, VQ-VAE latent vectors and the visual feature of layer i in CLIP extracted from the training set. Here, $D_j$ denotes the dimentions of aforementioned data,  and N denotes the size of the training set.
Figure \ref{fig:overview} (a) illustrates the training process of three linear regression models: $f_c:X \mapsto c$ , $f_z:X \mapsto z$ and $f_{CLIPi}:X \mapsto Z_{CLIP}^i$ using the training set. The trained $f_c$ and $f_z$ are utilized to decode CLIP text embeddings $c$ and latent vectors $z$ of the images in test set. Subsequently, these two feature vectors are fed into  Stable Diffusion, as illustrated in Figure \ref{fig:overview} (b).  To incorporate image prior into the latent space of diffusion model, decoded $z$ undergoes a forward diffusion process for 50 steps, as outlined by equations \ref{equ1} and \ref{equ2} , resulting in the computation of $z_T$.
\begin{equation}\label{equ1}
	q(z_t|z_{t-1})=\mathcal{N}(z_t;\sqrt{\alpha_t}z_{t-1},(1-\alpha_t)I)    \quad t=0,1,\cdots T,
\end{equation}
\begin{equation}\label{equ2}
	z_T=\sqrt{\overline{\alpha_T}} z+ \sqrt{1-\overline{\alpha_T}}\epsilon        \quad  and \quad z_0=z.
\end{equation}

In each reverse denoising iteration, the U-Net \cite{ronneberger2015u} integrates decoded CLIP text embedding $c$ into $z_T$ by cross-attention, as defined in Equation \ref{equ3}.
\begin{equation}
	\label{equ3}
	\begin{split}
		CrossAttention(Q,K,V)=softmax(\frac{Q K^T}{\sqrt{d}}), \\  Q=W_Q^i\cdot\phi_i(z_t), K=W_K^i\cdot c, V=W_V^i\cdot c.
	\end{split}
\end{equation}
where $\phi_i(z_t)$ represents the middle-layer feature of U-Net \cite{ronneberger2015u} , $c$ corresponds to the decoded CLIP text information, and $W_Q^i$, $W_K^i$, $W_V^i$ denote the pre-trained projection matrixs. In this way, we reformulate the optimization objective of Stable Diffusion to Equation \ref{equ4}.
\begin{equation}\label{equ4}
	L_{Semantic}^t= \mathbb{E} _{z_t,\epsilon\sim\mathcal{N}(0,1),t}[\Vert\epsilon-\epsilon_\theta(z_t,t,c,\phi_i(z_t)) \Vert_2^2].
\end{equation}
where $\epsilon_\theta( \cdot )$ is a set of denoising functions that are usually
implemented as U-Net. The images generated by this process contain semantic information and fine-grained detail.

\subsection{Stage 2: Structural information aligning} 

In stage 1, we employ the decoded CLIP text embedding $c$, and the VQ-VAE latent vector $z$ to generate an initial reconstructed image $\hat{Y}$ that contains semantic information. The CLIP visual branch, denoted as $\Phi$, encodes high-level semantic information at the last layer and lower-level structural information such as posture and position at the shallow layers, which is revealed experimentally by Wang et al. \cite{Wang2022.09.27.508760}. To align the structural information of the reconstructed image with that of the original one, we devise a structural loss function based on low-level features extracted from CLIP visual encoder:

\begin{equation}\label{equ5}
	L_{Structure}=\sum\limits_{i}\Vert \Phi^i_{CLIP}(\hat{Y})-Z^i_{CLIP} \Vert_2^2.
\end{equation}

As illustrated in Figure \ref{fig:overview} (b), by iteratively backpropagating the gradients of this loss function with respect to $c$ and $z$, we adjust the both components. Specifically, we begin by extracting low-level features from the preliminary image generated in the first stage using the CLIP visual encoder. Next, we compute the L2 norm between the extracted features and the structural features decoded from brain signals, which serves as our structural loss function. As all operations in the model are differentiable, we adjust the $c$ and $z$ by backpropagation. The updated $c$ and $z$ are then fed into the image generator of the first stage to update the reconstructed image. This process is repeated until convergence, achieving the goal of controlling the reconstructed output.

\section{Experiments}
\subsection{Dataset}

NSD \cite{allen2022massive} is currently the largest neuroimaging dataset bridging brain and artificial intelligence, consisting of densely sampled fMRI data from 8 subjects. Each subject viewed 9000-10000 different natural scenes (with 22000-30000 repetitions) during 30-40 MRI scanning sessions, using whole-brain gradient-echo EPI at 1.8 mm isotropic resolution and 1.6 s TR for 7T scanning. The image stimuli viewed by the subjects are obtained from the Common Objects in Context  (COCO) \cite{lin2014microsoft} dataset and corresponding captions can be extracted using the COCO ID of the image stimuli.

\begin{table}[htbp!]
	\centering
	\renewcommand\arraystretch{1.2}
	\scalebox{0.85}{
		\begin{tabular}{c|c|c|c|c|c}
			\toprule
			Dataset & Training & Test(shared) & ROIS & Subject ID & Voxels \\ \hline
			\multirow{4}{*}{NSD} & \multirow{4}{*}{8859} & \multirow{4}{*}{982} & \multirow{4}{*}{\begin{tabular}[c]{@{}c@{}}V1, V2, V3, hV4,\\ VO, PHC, MT,\\ MST, LO, IPS\end{tabular}} & sub01 & 11694 \\ 
			&          &              &      & sub02      & 9987   \\ 
			&          &              &      & sub05      & 9312   \\ 
			&          &              &      & sub07      & 8980   \\ \bottomrule
	\end{tabular}}
	\caption{The detail of NSD used in our experiments.}
	\label{tab:NSD}
\end{table}

To validate the stability of MindDiffuser across different subjects, we conducted experiments using fMRI-stimulus pairs from subjects 1, 2, 5, and 7 in NSD. The training set for each subject contains 8859 image stimuli and 24980 fMRI trials (with each image having up to 3 trials). Additionally, in the test set, 982 image stimuli and 2770 fMRI trials are shared among the four subjects. For fMRI data with multiple trials, we computed the average response. The properties of the dataset used in our experiments have been summarized in Table \ref{tab:NSD}.

\subsection{Feature decoding experiments}

During the image reconstruction process, we first utilize the L2-regularized Linear Regression\footnote{We employ \href{https://github.com/KamitaniLab/PyFastL2LiR}{FastL2LiR} in order to accomplish this task.} to decode fMRI data into three distinct spaces:

\begin{figure}[htbp!]
	\centering
	\includegraphics[width=1.0\linewidth]{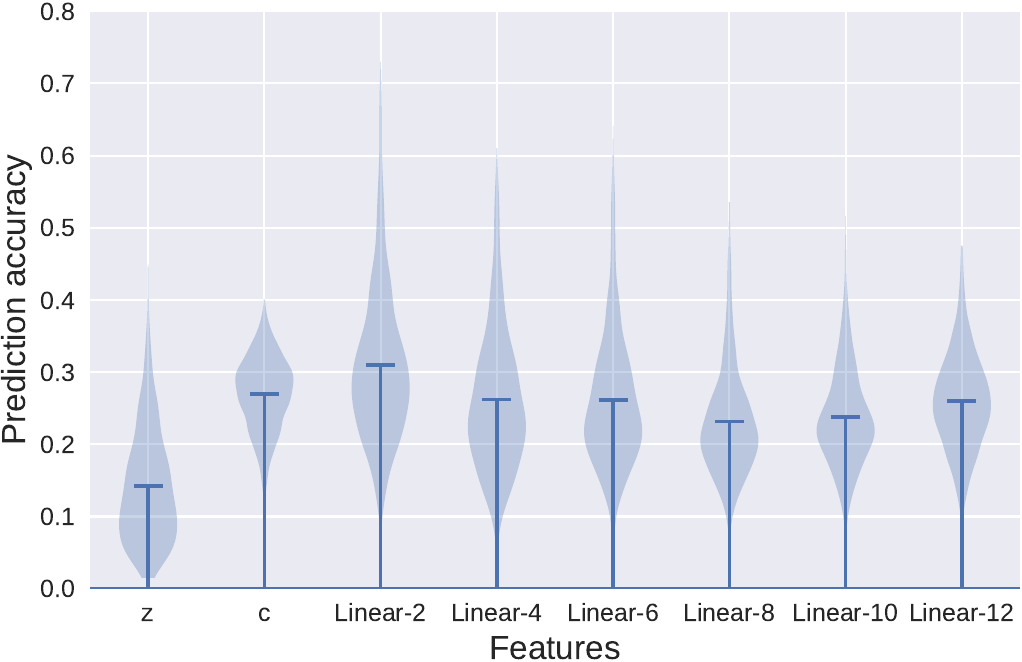}
	\caption{ Decoding accuracy of each feature. The results have
		been averaged over 4 subjects. The blue bars represent the average prediction
		accuracy of all units in each feature.}
	\label{fig:decodingaccuracystage1}
\end{figure}

{\bfseries CLIP text feature space.} The Stable Diffusion utilizes a CLIP text encoder with a feature space dimension of 77$\times$768, where 77 denotes the maximum token length and 768 represents the embedding dimension of each token. To account for the typical caption lengths in  COCO , which rarely exceed 15 tokens, the first 15$\times$768 dimensions of the flattened features are used during practical operation. The features in this space inject {\bfseries semantic } information into the reconstructed images.

{\bfseries VQ-VAE latent space.} To integrate richer {\bfseries detail}, we extract latent space feature (1$\times$4$\times$64$\times$64 dimensions) of the images in training set using VQ-VAE (included with Stable Diffusion) and flatten them. Subsequently, fMRI signals are mapped to this latent space. 

{\bfseries CLIP image feature space.} In order to align the {\bfseries structural} information of the reconstructed images with corresponding ground truth, we choose the low-level visual feature of CLIP to control the reconstructed images, as shown in Figure \ref{fig:overview} . We choose ViT/B-32 \cite{dosovitskiy2020image} as the backbone of pre-trained CLIP, and extract features from several layers (Linear-2, Linear-4, $\cdots$ Linear-12) in the CLIP visual branch. These layers have 38400 dimensions each. Due to the potential impact of those low-accuracy feature dimentions  in the decoding process, we first calculate the accuracy (Pearson correlation coefficient) of the CLIP features for each dimension in the training set using a 5-fold cross-validation. We then select the top k\% of features based on their predictive accuracy and re-fit these features using all the training data. During the reconstruction of images on the test set, only these k\% of features are used to guide the alignment of the original and reconstructed images. In this paper, we set k=25, and the methodology for selecting k is elaborated in Appendix \ref{appendix:select_k}.

The decoding accuracy\footnote{The computing process is partly implemented by MindSpore.} (Pearson correlation coefficient) of the three aforementioned features is presented in Figure \ref{fig:decodingaccuracystage1}.

\subsection{ Image reconstruction from decoded features}
{\bfseries Overview of our reconstruction results}. Following the process in Figure \ref{fig:overview} , some of the images reconstructed by MindDiffuser are shown in Figure \ref{fig:result-overview}. The upper bound illustrated in Figure \ref{fig:result-overview} evinces that our proposed MindDiffuser possesses an incredibly robust reconstruction capability, wherein it can faithfully reconstruct results that are almost indistinguishable from the original images, provided truth features are given.

\begin{figure}[htbp!]
	\centering
	\includegraphics[width=1.0\linewidth]{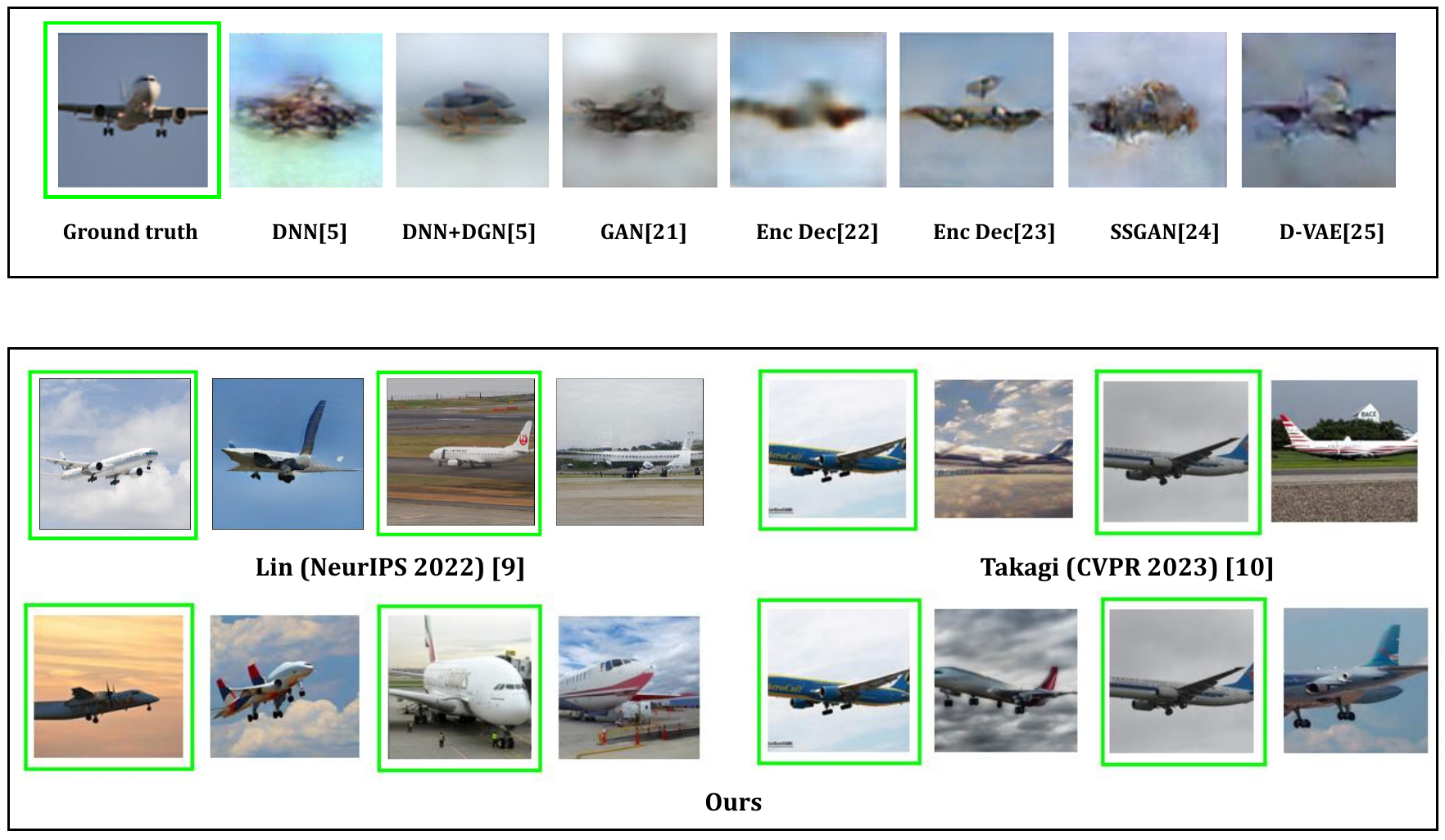}
	\caption{A brief comparison of image reconstruction results. }
	\label{fig:plane}
\end{figure}

Given that various reconstruction models employ distinct datasets, we opt to briefly compare results fairly by utilizing the aircraft reconstruction outcomes, which are prevalent across the majority of datasets and facilitate an intuitive comparison. Figure \ref{fig:plane} demonstrates that previous image reconstruction models, such as Shen DNN \cite{shen2019deep} and Shen GAN \cite{shen2019deep}, produce outcomes that resemble the original images in terms of size, shape, and orientation, but lack semantic information necessary for recognition as an airplane\footnote{ Note that the subfigure above shows the results of an previous image reconstruction collated by Lin \cite{lin2022mind}.}. Recent image reconstruction models, such as Mind Reader \cite{lin2022mind} and Takagi's \cite{takagi2022high}, incorporate text representations via multimodal pre-trained models, resulting in reconstructions with correct semantic information. However, in comparison to our MindDiffuser, their reconstructed aircrafts cannot be aligned with the original images on structural information such as shape and posture.

\begin{figure*}[htbp!]
	\centering
	\includegraphics[width=0.92\linewidth]{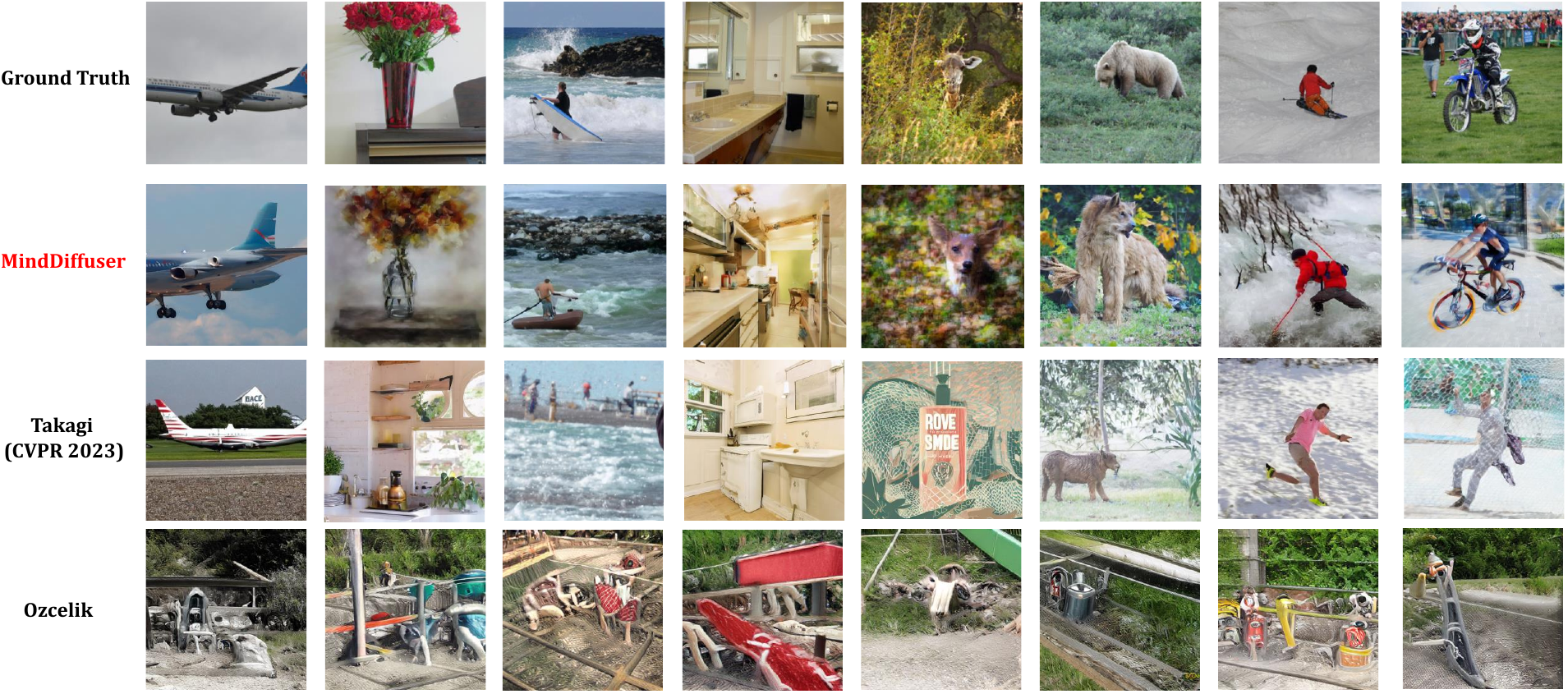}
	\caption{A comparative analysis of reconstruction models on NSD. The first line represents the image stimuli. The second line represents the results obtained using our proposed MindDiffuser. The third and fourth lines represent the results obtained by reproducing the experiment described in Takagi's \cite{takagi2022high} paper and by using Ozcelik's \cite{ozcelik2022reconstruction} provided code, respectively.}
	\label{fig:compare-with-sota}
\end{figure*}

\begin{figure*}[htbp!]
	\centering
	\includegraphics[width=0.95\linewidth]{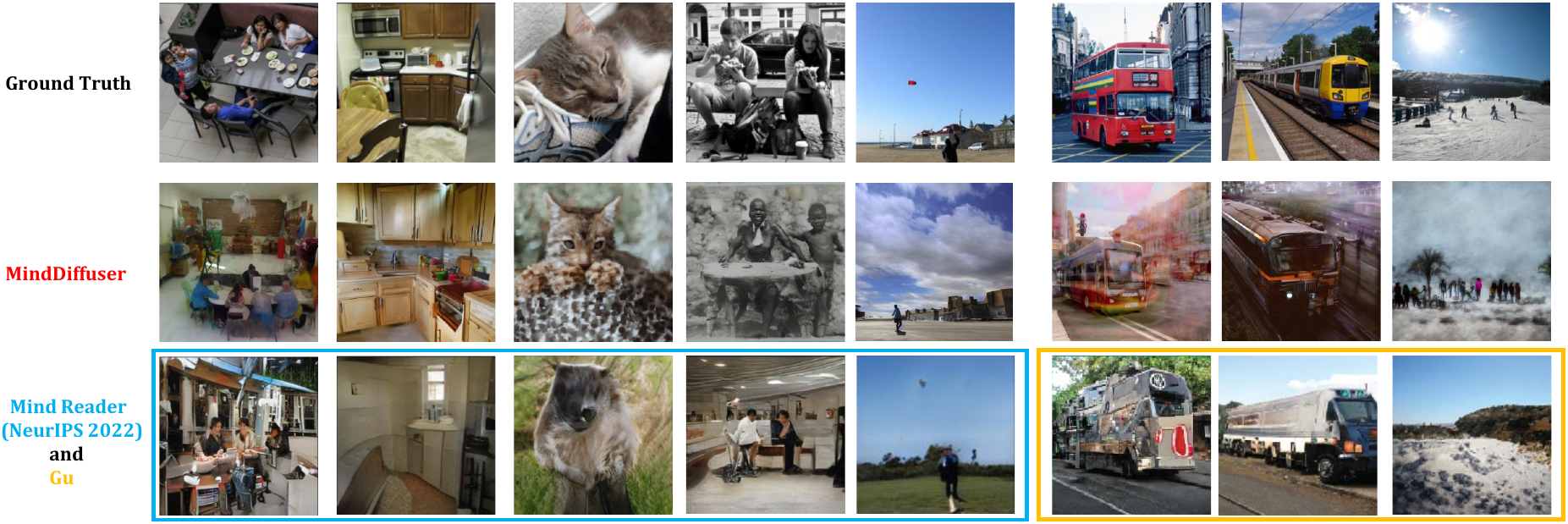}
	\caption{A comparative analysis of reconstruction models on NSD. The ground truth is presented in the first line, followed by our proposed method MindDiffuser in the second line. The reconstructed results shown in the blue box are from the work of Lin et al. \cite{lin2022mind}, and we used the same training and testing data partitioning for comparative experiments. The reconstructed results presented by Gu et al. \cite{gu2022decoding} are highlighted in a yellow box.}
	\label{fig:compare-with-lingu}
\end{figure*}

{\bfseries Comparison with state-of-the-art models}.As much of the existing work on image reconstruction utilizing NSD has not yet been made fully open source, we took it upon ourselves to reproduce the results of Takagi et al. \cite{takagi2022high} and Ozcelik et al. \cite{ozcelik2022reconstruction} on NSD, in order to facilitate direct comparison with our findings, as shown in Figure \ref{fig:compare-with-sota}. We conducted a comparative evaluation of our reconstruction outcomes with the results demonstrated by Lin \cite{lin2022mind} and Gu \cite{gu2022decoding} in their published research, as delineated in Figure \ref{fig:compare-with-lingu}.

\begin{table}[hbp!]
	\centering
	\renewcommand\arraystretch{1.2}
	\scalebox{0.85}{
		\begin{tabular}{c|c|cc}
			\toprule
			\multirow{2}{*}{Methods} & Semantic Similarity {\bfseries $\uparrow$} & \multicolumn{2}{c}{Structure Similarity}   {\bfseries $\uparrow$}          \\ \cline{2-4} 
			& CLIP Similarity               & \multicolumn{1}{c|}{SSIM}           & PCC            \\ \hline
			\multirow{4}{*}{\begin{tabular}[c]{@{}c@{}}Ozcelik \cite{ozcelik2022reconstruction}\\ Takagi (CVPR2023) \cite{takagi2022high}\\ Lin (NeurIPS 2022) \cite{lin2022mind}\\ Gu \cite{gu2022decoding}\end{tabular}} &
			\multirow{4}{*}{\begin{tabular}[c]{@{}c@{}}0.546\\ 0.642\\ 0.578\\ 0.721\end{tabular}} &
			\multicolumn{1}{c|}{\multirow{4}{*}{\begin{tabular}[c]{@{}c@{}}0.135\\ 0.300\\ 0.243\\ 0.117\end{tabular}}} &
			\multirow{4}{*}{\begin{tabular}[c]{@{}c@{}}0.126\\ 0.175\\ 0.102\\ 0.188\end{tabular}} \\
			&                     & \multicolumn{1}{c|}{}               &                \\
			&                     & \multicolumn{1}{c|}{}               &                \\
			&                     & \multicolumn{1}{c|}{}               &                \\ \hline
			Ours                     & \textbf{0.765}      & \multicolumn{1}{c|}{\textbf{0.354}} & \textbf{0.278} \\ \bottomrule 
	\end{tabular}     }
	\caption{Quantitative comparison of image reconstruction. Three metrics are utilized to evaluate the semantic and structural similarity between the original and reconstructed images. Larger metric values indicate better reconstruction quality, with the best results emphasized in bold. }
	\label{tab:all_part}
\end{table}

According to Figures \ref{fig:compare-with-sota} and \ref{fig:compare-with-lingu}, it can be observed that compared to recent work, our approach produces reconstructed results on NSD that are visually more similar to the ground truth both in terms of semantic and structure. In order to further quantitatively compare the reconstruction performance of our method with some state-of-the-art methods, we utilize three evaluation metrics to compare from both semantic and structural aspects. We use cosine similarity in the last layer of CLIP visual branch (the dimension is 512) to measure the semantic similarity between the reconstructed results and the original images. We use SSIM and per-pixel correlation coefficient (PCC) to measure the structural similarity between them. 

All the three metrics range from 0 to 1, and higher values indicate better reconstruction results. For the convenience of comparison, the  evaluation metrics (CLIP, SSIM, PCC) are calculated based on the reconstructed results (resized to 512 $\times$ 512) shown in Figures \ref{fig:compare-with-sota} and \ref{fig:compare-with-lingu}. The calculation results on the whole test set can be found in Table \ref{tab:test set} in the appendix. The results from Tables \ref{tab:all_part}  and \ref{tab:test set} demonstrate that our MindDiffuer surpasses the current state-of-the-art models both semantically and structurally.

\begin{figure}[htbp!]
	\centering
	\includegraphics[width=1.0\linewidth]{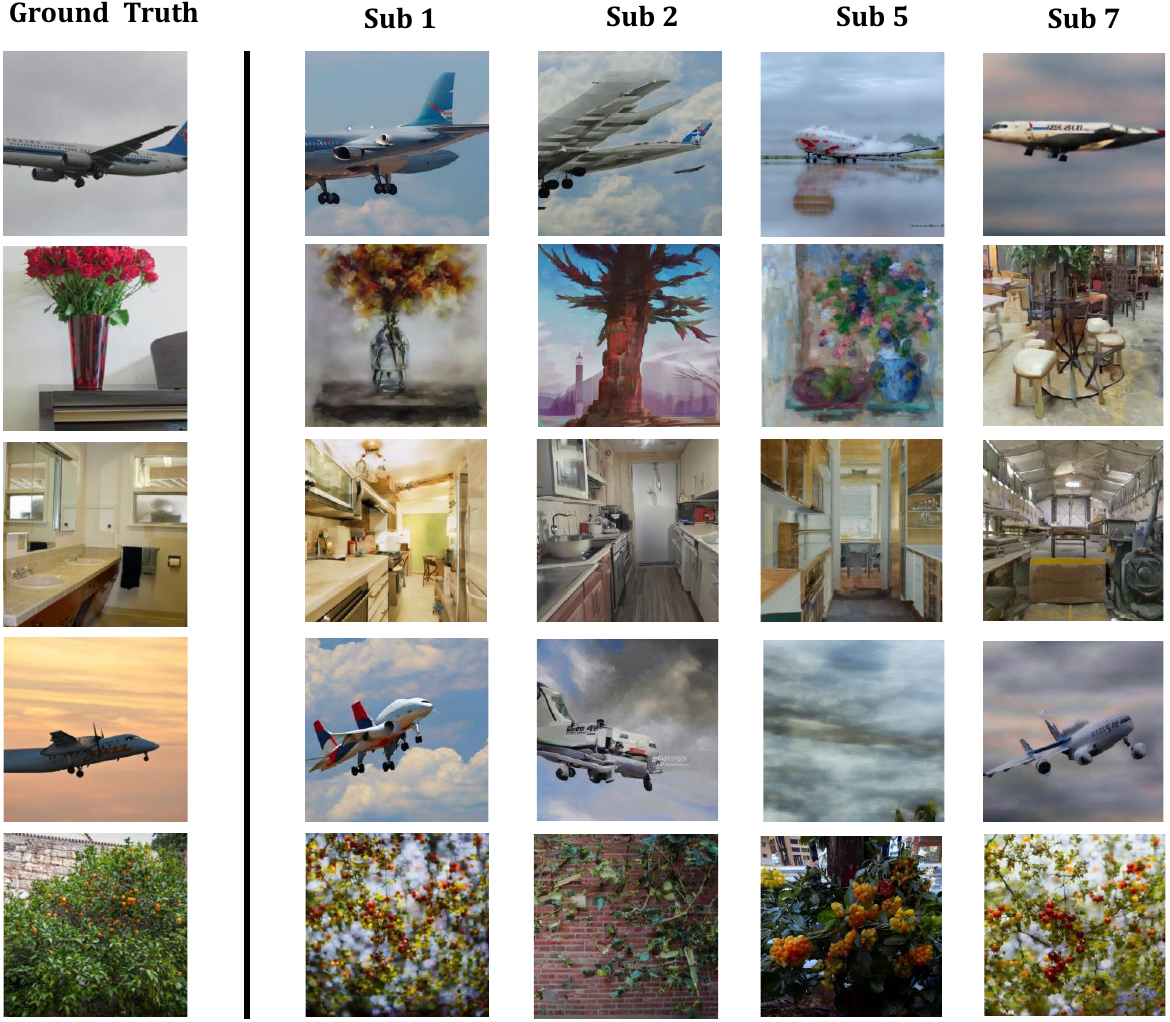}
	\caption{Reconstruction results of MindDiffuser on multiple subjects}
	\label{fig:foursub}
\end{figure}

{\bfseries Adaptability across 4 subjects}. The anatomical structure and functional connectivity of the brain vary among individuals \cite{cohen2008defining}, resulting in differences in the fMRI signals even when the same image stimulus is presented. To validate MindDiffuser's ability to adapt to inter-subject variability, we reconstruct the test images of subjects 1, 2, 5, and 7 without any additional adjustments. The results are shown in Figure \ref{fig:foursub}.

As depicted in Figure \ref{fig:foursub}, the identical image stimulus may result in different and unsatisfactory reconstruction results for some subjects due to differences in the subjective brain responses during fMRI acquisition and disparities in the accuracy of feature decoding. For instance, "flowers on the table" is erroneously reconstructed as "table and chair" in subject 7, and "aircraft at sunset" can not be reconstructed in subject 5. However, for the majority of the reconstructed images, our model achieves a satisfactory alignment with the original images in terms of both semantic and structural features for each subject, underscoring the ability of MindDiffuser to effectively accommodate variations among individuals.

\subsection{Ablation study}

\begin{figure}[htbp!]
	\centering
	\includegraphics[width=1.0\linewidth]{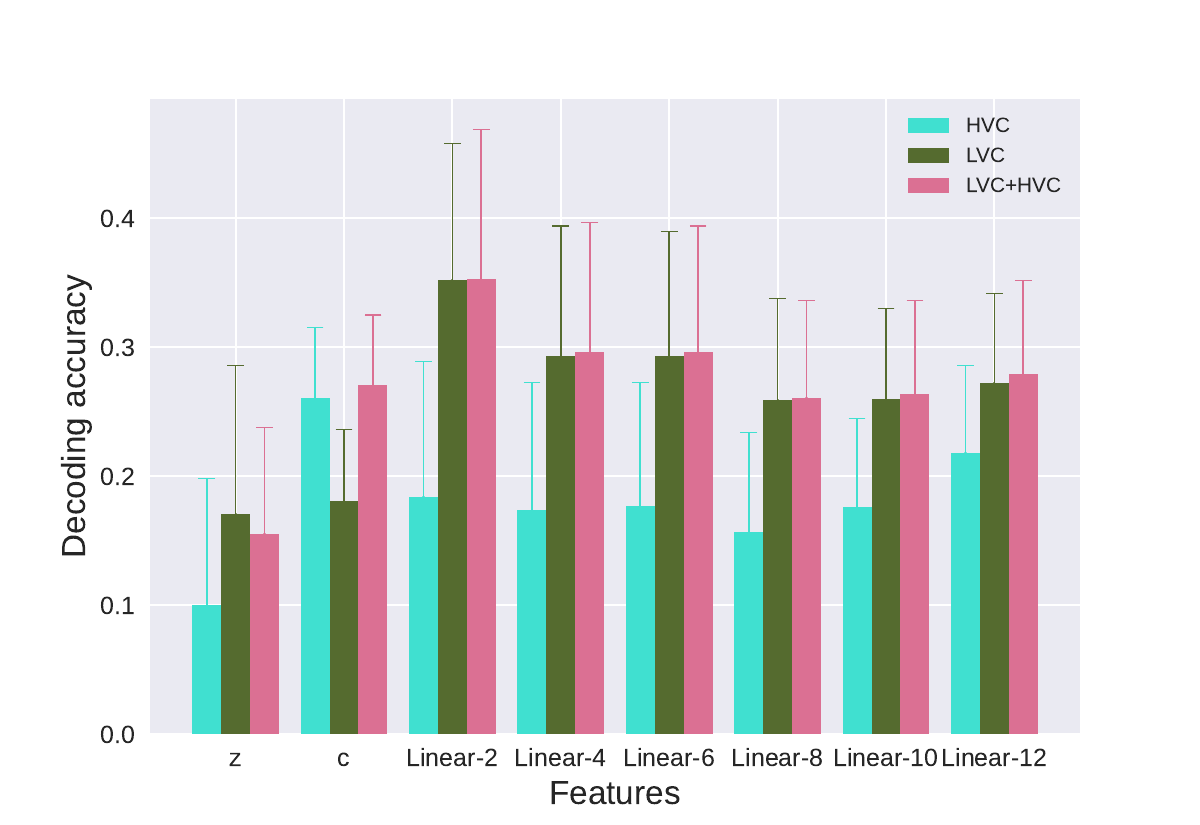}
	\caption{The decoding accuracy of different ROIs during the decoding phase. All experiments are conducted on subject 1.}
	\label{fig:ablationforvc}
\end{figure}

During the feature decoding stage, we utilize the voxels from both Low-level Visual Cortex (LVC) and High-level Visual Cortex (HVC) provided in NSD. The LVC demonstrates a preference for responding to local visual information, such as shape and texture, while the HVC integrates visual elements and information over a larger receptive field, enabling the perception of global visual stimuli and semantic cognition \cite{felleman1991distributed, gazzaley2012top}. To investigate the effect of information contained within these ROIs on reconstruction results, we separately use voxels from either LVC or HVC during the decoding process. Results from the first ablation experiment are shown in Figure \ref{fig:ablationforvc} and Table \ref{tab:fMRI_ablation}. 

\begin{table}[htbp!]
	\centering
	\renewcommand\arraystretch{1.2}
	
	\begin{tabular}{ll|c|cc}
		\toprule
		\multicolumn{2}{c|}{Visual cortex } & \multicolumn{1}{l|}{Semantic similarity {\bfseries $\uparrow$}} & \multicolumn{2}{l}{Structual similarity {\bfseries $\uparrow$}}             \\ \hline
		\multicolumn{2}{c|}{LVC  \quad   HVC} & CLIP Similarity  & \multicolumn{1}{c|}{SSIM}  & PCC   \\ \hline
		\multicolumn{2}{l|}{\ding{51}}           & 0.552 & \multicolumn{1}{c|}{0.338} & 0.051 \\ 
		\multicolumn{2}{l|}{ \qquad \qquad\ding{51}}           & 0.554 & \multicolumn{1}{c|}{0.219} & 0.030  \\ 
		\multicolumn{2}{l|}{\ding{51}\qquad \,\quad\ding{51}}     & \textbf{0.765}                           & \multicolumn{1}{c|}{\textbf{0.354}} & \textbf{0.278} \\ \bottomrule 
	\end{tabular}
	\caption{Quantitative comparison of ablation experiment results on different ROIs. The best results are emphasized in bold.}
	\label{tab:fMRI_ablation}
\end{table}

\begin{table}[htbp!]
	\centering
	\renewcommand\arraystretch{1.1}
	\scalebox{0.85}{
		\begin{tabular}{l|c|c|cc}
			\toprule
			\multicolumn{1}{c|}{\multirow{2}{*}{\textbf{Model}}} &
			\multicolumn{1}{c|}{\multirow{2}{*}{\textbf{Variants}}} &
			\multicolumn{1}{l|}{Semantic similarity {\bfseries $\uparrow$}} &
			\multicolumn{2}{l}{Structual similarity {\bfseries $\uparrow$}} \\ \cline{3-5} 
			\multicolumn{1}{c|}{} & \multicolumn{1}{l|}{}   & CLIP Similarity           & \multicolumn{1}{c|}{SSIM}           & PCC            \\ \hline
			MindDiffuser          & w/o     $c$     & 0.549          & \multicolumn{1}{c|}{0.346}          & 0.218          \\ 
			MindDiffuser          & w/o    $z$       & 0.616          & \multicolumn{1}{c|}{0.292}          & 0.066          \\ 
			MindDiffuser          & w/o     $Z_{CLIP}$ & 0.597          & \multicolumn{1}{c|}{0.253}          & 0.183          \\ 
			MindDiffuser    & completed               & \textbf{0.765} & \multicolumn{1}{c|}{\textbf{0.354}} & \textbf{0.278} \\ \bottomrule 
	\end{tabular}}
	\caption{Quantitative comparison of ablation experiment results on different features. The best results are emphasized in bold.}
	\label{tab:zcclip}
\end{table}
We note that LVC is more important for decoding structural and detailed information ( $Z_{CLIP}$ and $z$ ), while HVC is more important for decoding semantic information ( $c$ ), which is consistent with findings in neuroscience.

Next, we conduct ablation experiments on the three features used in MindDiffuser: semantic feature $c$, detail feature $z$, and structural feature $Z_{CLIP}$. The results, shown in Table \ref{tab:zcclip}, demonstrate that MindDiffuser outperforms other variants in reconstruction. Removing any one of the features significantly reduces the reconstruction performance. 

\subsection{Neuroscience interpretability}
\begin{figure}[htbp!]
	\centering
	\includegraphics[width=1.0\linewidth]{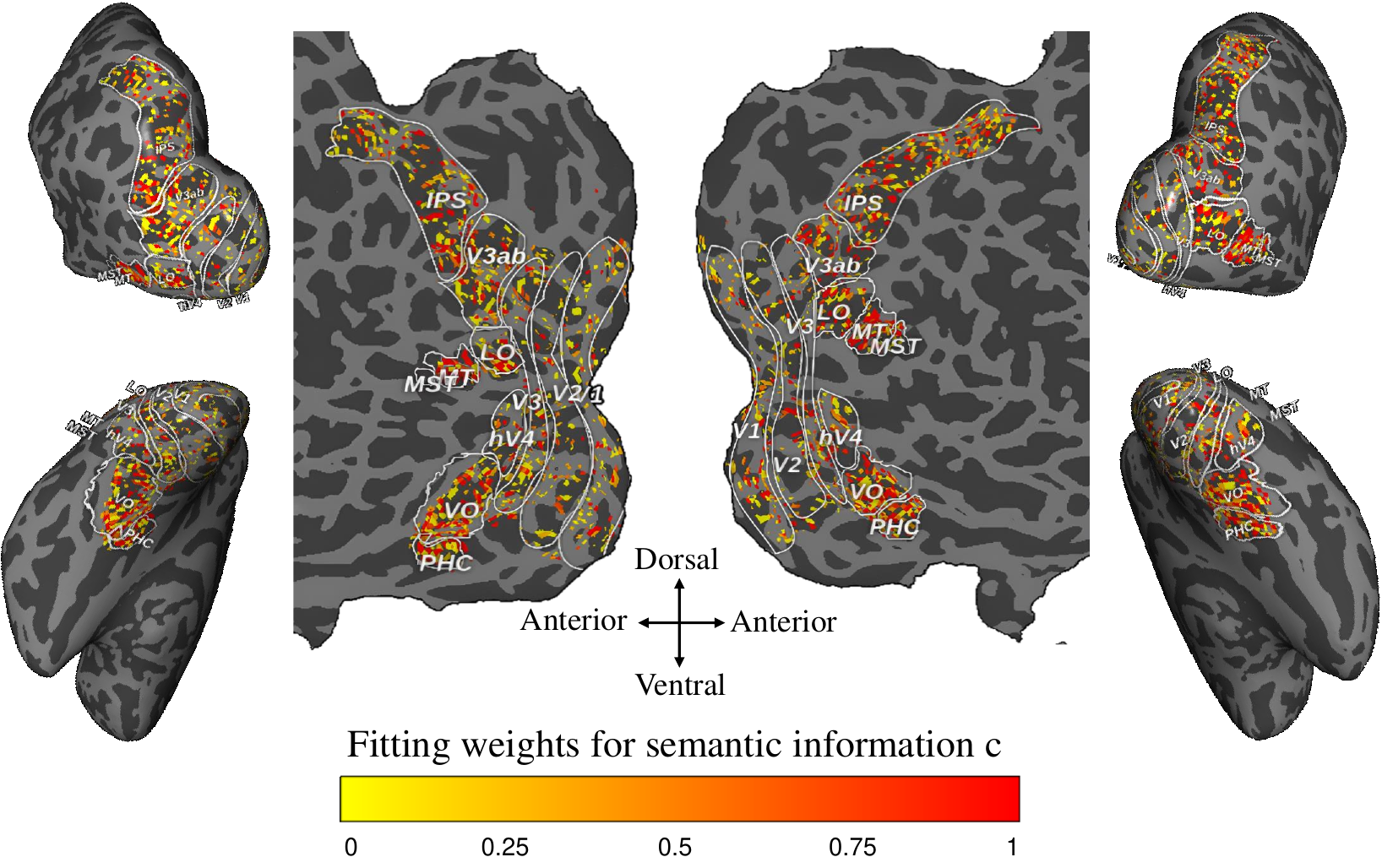}
	\caption{The importance of different ROIs of subject 1 in decoding the semantic feature $c$. The weights of  trained linear regression model are projected onto the corresponding locations of each voxel on the cortical surface. We flatten the cortical surface into a 2D plane and discard the points with zero weights. The redder area in the figure indicates a higher weight in decoding.}
	\label{fig:cortexc}
\end{figure}

\begin{figure}
	\centering
	\includegraphics[width=1.0\linewidth]{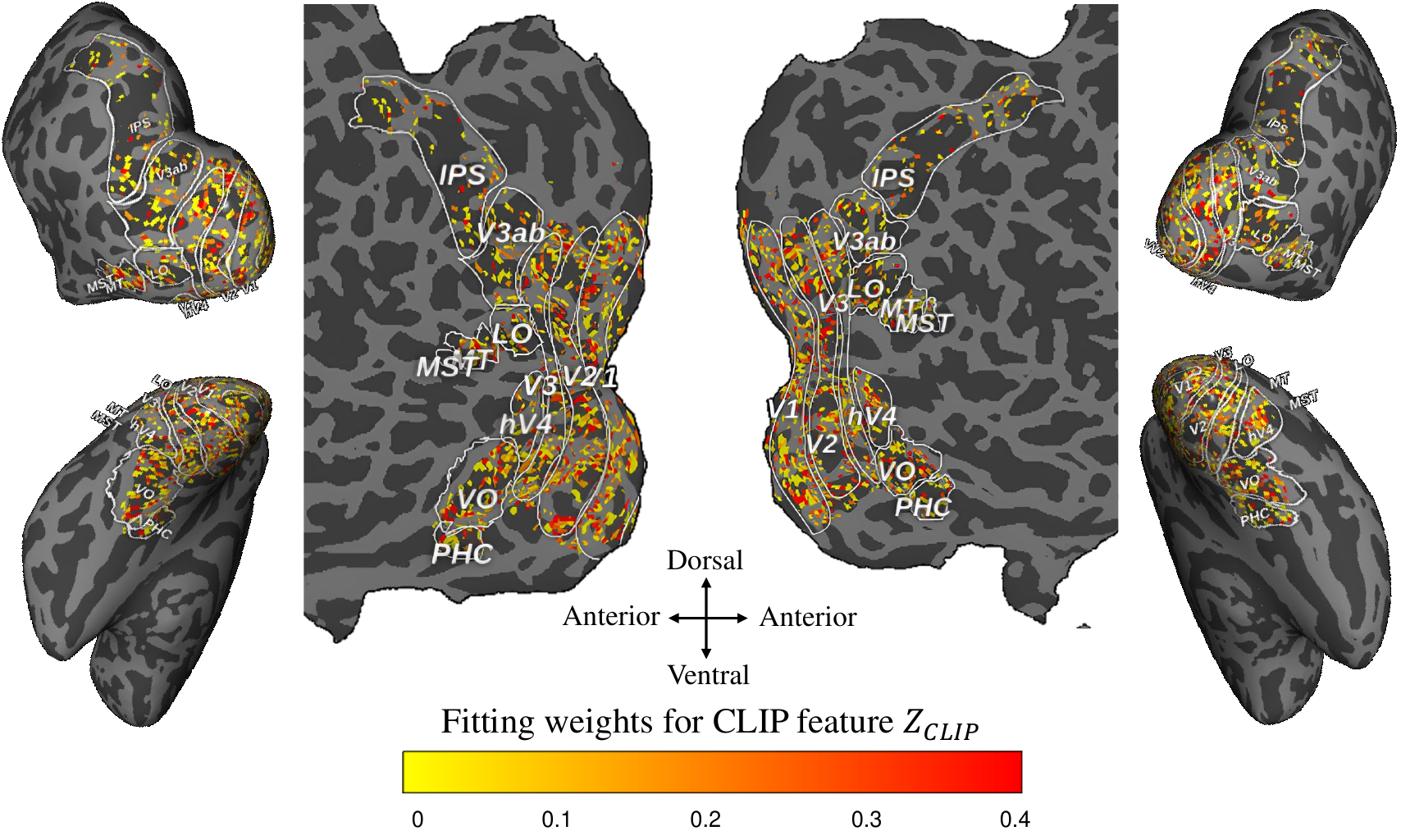}
	\caption{The importance of different ROIs of subject 1 in decoding the structural feature $Z_{CLIP}$.}
	\label{fig:cortexclip}
\end{figure}

\begin{figure}[htbp!]
	\centering
	\includegraphics[width=1.0\linewidth]{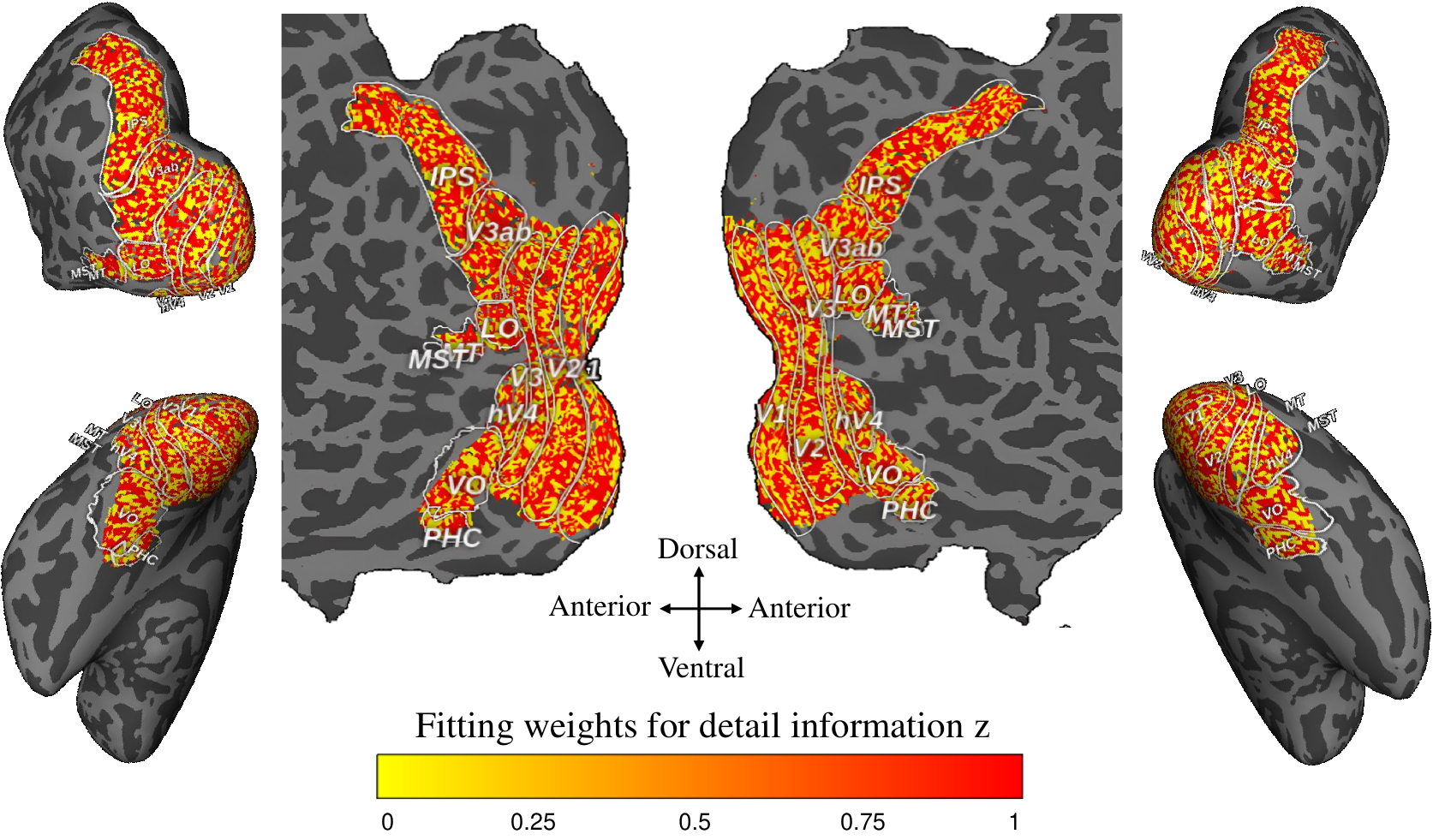}
	\caption{The importance of different ROIs of subject 1 in decoding the detail feature $z$.}
	\label{fig:cortexz}
\end{figure}

To investigate the interpretability of MindDiffuser in neuroscience, we conduct the following experiments. During the feature decoding process, we use L2-regularized linear regression model  to automatically select voxels to fit three types of feature: semantic feature $c$, detail feature $z$, and structural feature $Z_{CLIP}$. We ultilize pycortex \cite{10.3389/fninf.2015.00023} to project the weights of each voxel in the fitted model onto the corresponding 3D coordinates in the visual cortex\footnote{\url{https://github.com/gallantlab/pycortex}}, as shown in Figures \ref{fig:cortexc}, \ref{fig:cortexz},  \ref{fig:cortexclip}. 
\hspace*{\fill}
\hspace*{\fill}
\hspace*{\fill}
\hspace*{\fill}

The visualization from Figure \ref{fig:cortexc} shows that when decoding semantic feature $c$, most of the selected voxels are located in IPS, LO, MT, MST, PHC, and VO that process high-level semantic information \cite{Wang2022.09.27.508760}. Moreover, the distribution of voxel weights within the high-level visual cortex seems to be noticeably more pronounced than that within the low-level visual cortex.
Figure \ref{fig:cortexclip} reveals that the CLIP features $Z_{CLIP}$ utilized for supervising the reconstruction of image structure in Stage 2 are primarily fitted by low-level visual cortex, such as V1, V2, V3, V3ab, and hV4 \cite{Wang2022.09.27.508760}, which process local shape and texture information. We observe that in our brain decoding process, semantic feature $c$ and structural feature $Z_{CLIP}$ are mainly explained by high-level and low-level visual cortex seperately, which is consistent with the antecedent endeavors of the predecessors \cite{pinto2009high, krizhevsky2009learning, schrimpf2018brain, Wang2022.09.27.508760}.

Figure \ref{fig:cortexz} shows that voxels from both high-level and low-level visual cortex are involved in decoding detail feature $z$, with comparable weighting magnitudes. This indicates that in stage 1, decoded $z$  enables the cross-attention mechanism to incorporate fine-grained semantic and structural detail into the reconstructed images.

The aforementioned findings corroborate the neurobiological plausibility of our model, as evidenced by the interpretability of the multimodal features employed, which align with the corresponding brain responses.
\hspace*{\fill}
\section{Limitations and future work}
Due to the limited temporal resolution of fMRI signals, our image reconstruction model, while effective for static images, faces challenges when directly applied to video reconstruction tasks. To address this issue, we aim to explore temporal channel modeling methods to reconstruct motion visual signals from brain. This provides a promising avenue for the reconstruction of controlled and realistic videos from brain in the  future.
\hspace*{\fill}
\section{Conclusion}

We propose a two-stage image reconstruction model, {\bfseries MindDiffuser}, which aligns both semantic and structural information of the reconstructed images with the image stimuli.  MindDiffuser outperforms the current state-of-the-art models qualitatively and quantitatively on NSD, given the high decoding accuracy of stage 1. Moreover, our experiments show that MindDiffuser is adaptive to inter-subject variability, achieving excellent reconstruction results for the stimuli of subjects 1, 2, 5, and 7 without any additional adjustment. Furthermore, the experiments demonstrate that the multimodal information utilized in our model can be explained by corresponding brain responses in neuroscience, thereby providing validation for the rationality of the model design. We believe that  MindDiffuser plays a significant role in facilitating precise and controlled stimuli reconstruction for brain-computer interfaces.

\hspace*{\fill}
\begin{acks}
	This work was supported in part by the National Key R\&D Program of China 2022ZD0116500; in part by the National Natural Science Foundation of China under Grant 62206284, Grant 62020106015 and Grant 61976209;  and in part by the CAAI-Huawei MindSpore Open Fund.
\end{acks}
\hspace*{\fill}
\hspace*{\fill}

\balance
\bibliographystyle{ACM-Reference-Format}
\bibliography{sample-base}

\clearpage

\appendix

\nobalance

\section{Upper bound of the model's reconstruction ability}
\label{appendix:upperbound}
Our work consists of two parts. The first part involves training brain signal decoders (top left corner of Figure \ref{fig:overview}) that utilizes regression models to map brain signals to image features, such as semantic and structural features. The second part presents our innovative two-stage image reconstruction model, which does not require training. This model leverages the input image features to reconstruct the corresponding original image. Due to the impact of the accuracy of brain signal decoding on the image reconstruction quality, we directly input the real semantic ($c$) and structural ($z_{CLIP}$) features of the original images (instead of the features decoded from brain signals) into the model, considering the resulting images as the upper bound of the model's reconstruction ability.

\section{The selection of k for CLIP features }
\label{appendix:select_k}
\begin{figure}[htbp!]
	\centering
	\includegraphics[width=0.7\linewidth]{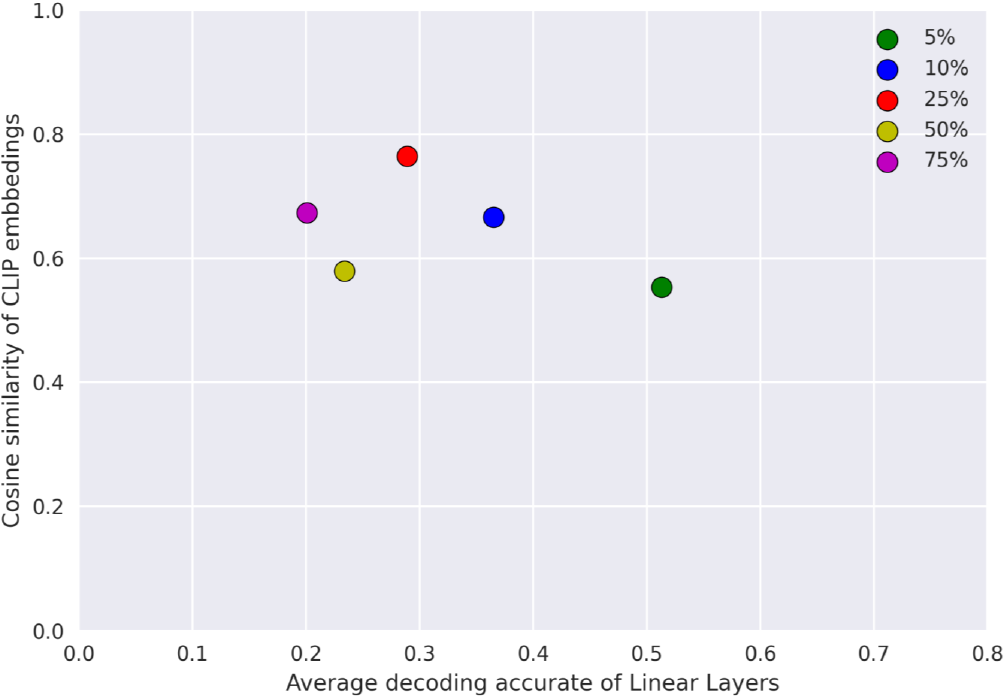}
	\caption{Illustration depicting the selection of retained percentage "k" for preserving CLIP features.  (CLIP)}
	\label{fig:selectk1}
\end{figure}
\begin{figure}[htbp!]
	\centering
	\includegraphics[width=0.7\linewidth]{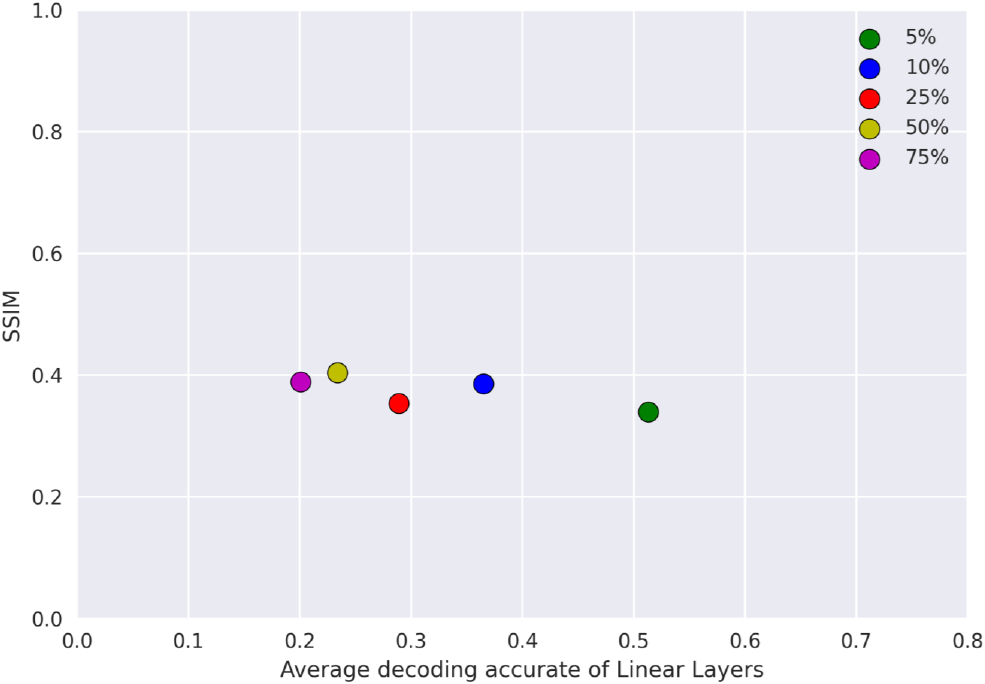}
	\caption{Illustration depicting the selection of retained percentage "k" for preserving CLIP features. (SSIM)}
	\label{fig:selectk2}
\end{figure}

\begin{figure}[htbp!]
	\centering
	\includegraphics[width=0.7\linewidth]{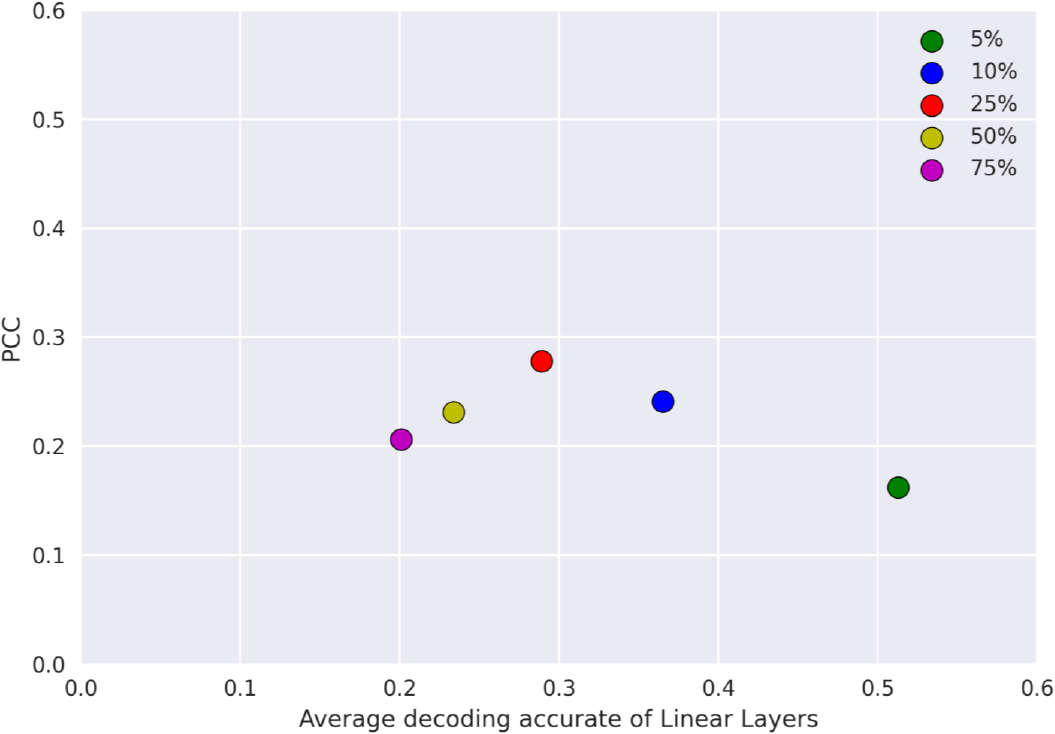}
	\caption{Illustration depicting the selection of retained percentage "k" for preserving CLIP features. (PCC)}
	\label{fig:selectk3}
\end{figure}
During feature decoding process, only those CLIP features $Z_{CLIP}$ with prediction accuracy in the top k\% are fitted. Hence, a smaller value of k leads to higher decoding accuracy, but also results in more information loss. Therefore, careful consideration is required to balance decoding accuracy and retained information when selecting the value of k. We evaluate different values of k (k=5, 10, 25, 50, and 75) using three metrics - CLIP embedding similarity, SSIM, and PCC - to measure the semantic and structural similarity between reconstructed and original images. The evaluation is performed on the validation set (last 859 images of the training set), and the results are shown in Figures \ref{fig:selectk1}, \ref{fig:selectk2}, \ref{fig:selectk3}.

Based on the scatter plots, points closer to the top-right corner indicate higher decoding accuracy and better preservation of semantic and structural information. When k=25, the reconstructed results on the validation set show the maximum values for CLIP embedding similarity and PCC, with SSIM results similar to other settings. Thus, we choose k=25 in subsequent experiments.

\section{Quantitative comparison results on the whole test set}
\label{appendix:whole test set}

In order to comprehensively and impartially evaluate the reconstruction performance of MindDiffuser on the NSD test set, we calculate the three metrics mentioned earlier as well as Fréchet Inception Distance (FID) \footnote{\url{https://github.com/mseitzer/pytorch-fid}} across the whole test set, as shown in Table \ref{tab:test set}.

\begin{table}[hbp!]
	\centering
	\renewcommand\arraystretch{1.0}
	\scalebox{1.0}{
		\begin{tabular}{c|c|cc|c}
			\toprule
			\multirow{2}{*}{Methods} & Semantic  {\bfseries $\uparrow$} & \multicolumn{2}{c|}{Structure  {\bfseries $\uparrow$}} & \multirow{2}{*}{FID {\bfseries $\downarrow$}} \\ \cline{2-4}
			& CLIP Similarity               & \multicolumn{1}{c|}{SSIM}      & PCC      &                      \\ \hline
			\multirow{2}{*}{\begin{tabular}[c]{@{}c@{}}Ozcelik \cite{ozcelik2022reconstruction}\\ Takagis  \cite{takagi2022high}\end{tabular}} &
			\multirow{2}{*}{\begin{tabular}[c]{@{}c@{}}0.549\\ 0.575\end{tabular}} &
			\multicolumn{1}{c|}{\multirow{2}{*}{\begin{tabular}[c]{@{}c@{}}0.245\\ {\bfseries0.246}\end{tabular}}} &
			\multirow{2}{*}{\begin{tabular}[c]{@{}c@{}}0.006\\ 0.186\end{tabular}} &
			\multirow{2}{*}{\begin{tabular}[c]{@{}c@{}}164.48\\ 127.69\end{tabular}} \\
			&                     & \multicolumn{1}{c|}{}          &          &                      \\ \hline
			\multirow{4}{*}{\begin{tabular}[c]{@{}c@{}}Ours (sub01)\\ Ours (sub02)\\ Ours (sub05)\\ Ours (sub07)\end{tabular}} &
			\multirow{4}{*}{\begin{tabular}[c]{@{}c@{}} {\bfseries0.603} \\ 0.588\\ 0.594\\ 0.579\end{tabular}} &
			\multicolumn{1}{c|}{\multirow{4}{*}{\begin{tabular}[c]{@{}c@{}}0.238\\ 0.249\\ 0.349 \\ 0.237\end{tabular}}} &
			\multirow{4}{*}{\begin{tabular}[c]{@{}c@{}} {\bfseries0.241}\\ 0.206\\ 0.150\\ 0.161\end{tabular}} &
			\multirow{4}{*}{\begin{tabular}[c]{@{}c@{}} {\bfseries112.83}\\ 120.19\\ 116.39\\ 121.06\end{tabular}} \\
			&                     & \multicolumn{1}{c|}{}          &          &                      \\
			&                     & \multicolumn{1}{c|}{}          &          &                      \\
			&                     & \multicolumn{1}{c|}{}          &          &                      \\ \hline
			Mean                    & 0.591               & \multicolumn{1}{c|}{0.268}     & 0.190     & 117.62               \\ \bottomrule
	\end{tabular}}
	\caption{Quantitative comparison results on the whole test set. The best results on subject 1 are emphasized in bold. }
	\label{tab:test set}
\end{table}

\end{document}